\documentclass[journal]{IEEEtran}
\usepackage{amsmath,amsfonts}
\usepackage{algorithm}
\usepackage{algpseudocode}
\usepackage{array}
\usepackage[caption=false,font=normalsize,labelfont=sf,textfont=sf]{subfig}
\usepackage{textcomp}
\usepackage{stfloats}
\usepackage{url}
\usepackage{verbatim}
\usepackage{graphicx}
\usepackage{cite}
\begin{document}

\title{Learning Trustworthy Model from Noisy Labels based on Rough Set for Surface Defect Detection}

\author{Tongzhi Niu, Bin Li, Kai Li, Yufeng Lin, Yuwei Li, Weifeng Li, Zhenrong Wang}

\markboth{Learning form noisy label for Surface Defect Inspection}%
{Shell \MakeLowercase{\textit{et al.}}: A Sample Article Using IEEEtran.cls for IEEE Journals}


\maketitle

\begin{abstract}
In the surface defect detection, there are some suspicious regions that cannot be uniquely classified as abnormal or normal. The annotating of suspicious regions is easily affected by factors such as workers' emotional fluctuations and judgment standard, resulting in noisy labels, which in turn leads to missing and false detections, and ultimately leads to inconsistent judgments of product quality. Unlike the usual noisy labels, the ones used for surface defect detection appear to be inconsistent rather than mislabeled. The noise occurs in almost every label and is difficult to correct or evaluate. In this paper, we proposed a framework that learns trustworthy models from noisy labels for surface defect defection. At first, to avoid the negative impact of noisy labels on the model, we represent the suspicious regions with consistent and precise elements at the pixel-level and redesign the loss function. Secondly, without changing network structure and adding any extra labels, pluggable spatially correlated Bayesian module is proposed. Finally, the defect discrimination confidence is proposed to measure the uncertainty, with which anomalies can be identified as defects. Our results indicate not only the effectiveness of the proposed method in learning from noisy labels, but also robustness and real-time performance.
\end{abstract}


\section{Introduction}
\IEEEPARstart{A}{S} an essential step in the manufacturing process, surface defect inspection is widely used in various industrial fields, including semiconductor electronics, automotive, pharmaceutical, chemical and other industries. Recently, deep convolutional neural networks have achieved impressive performance in surface defect detection \cite{Ren2022Data,Qu2022A,Massoli2022MOCCA,Dong2020PGA}. However, as a kind of data-driven model, deep learning models have been found to reproduce or amplify human errors and biases introduced into the training dataset during data labeling process \cite{Rich2019Lessons,Liao2022Learning, Xu2022Anti}. 

There are two main challenges in the annotation task: 1) objectively, there are some suspicious regions that cannot be uniquely classified as abnormal or normal, such as weak features and border regions. In many cases, the defect and background are the same material, with very similar colors and textures. The boundary between the defect and the background is usually not an absolute line, but a region. 2) subjectively, the labeling of weak feature and border regions are easily affected by factors including workers' unstable emotions, judgment standards and technique levels, resulting in noisy labels.

The suspicious regions are indistinguishable, which will generate noisy labels. Specifically, some samples are over-labeled, where the suspicious regions are annotated as anomalies. And some samples are under-labeled, where the suspicious regions are annotated as normal. Suspicious regions with same characteristics may be annotated as anomalies in some samples and in some other samples be annotated as normal. The noisy labels for surface detect detection are mainly characterized by inconsistency. 

\begin{figure}[!t]
  \centering
  \includegraphics[width=3.4in]{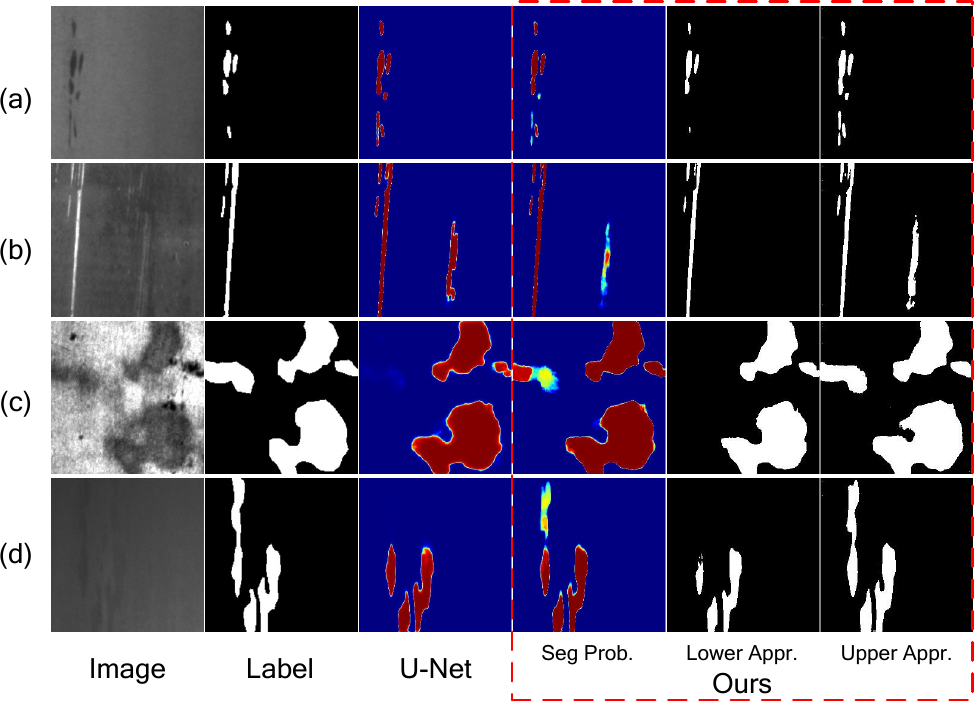}
  \caption{Results of our model learning from noisy labels. As can be seen from row (a) and (b), the suspicious regions are annotated as normal in the labels, but are falsely detected as abnormal in the U-net. In contrast, as shown in the row (c) and (d), the suspicious regions are marked as abnormal but detected as normal. In our methods, U-net based BNNs, suspicious regions are represented by segmentation probabilities. The lower approximations are same as the labels in false detections, and the upper approximations are same as the labels in missing detections.}
  \label{Fig. 1}
  \end{figure}
Due to the inconsistency of noisy labels, it is difficult to learn robust representations for suspicious regions. As shown in Fig. 1, the model learned from noisy labels performs as follows: 1) False detection. The model overfits the characteristics of suspicious regions that are over-labeled in the noisy labels. Some suspicious regions are detected as anomalies even though these regions are annotated as normal in the labels, as shown in row (a) and (b). 2) Missing detection. In contrast, some suspicious regions are detected as normal due to model underfitting, even though they are annotated as anomalies in the labels, as shown in row (c) and (d). The geometric dimensions of the abnormal regions (such as the length or diameter) are necessary indicators to judge whether the abnormality is defective. False and missing detections will lead to inaccurate measurement of the geometric dimensions of abnormal regions, resulting in inconsistent judgment of product quality. 

In order to learn a trustworthy discriminative model from noisy labels with inconsistencies, we focus on the following three aspects: 
1) to avoid the negative effects of noisy labels, consistent and precise elements in noisy labels are explored at pixel-level, and loss function is redesigned; 
2) without adding any extra labeling information and changing neural networks structure, pluggable spatially Bayesian modeule (PSBM) is designed to solve the suspicious regions of noisy labels; 
3) a discriminant confidence is proposed to measure the uncertainty of discriminating abnormality as a defect.

\begin{figure}[!t]
  \centering
  \includegraphics[width=3.4in]{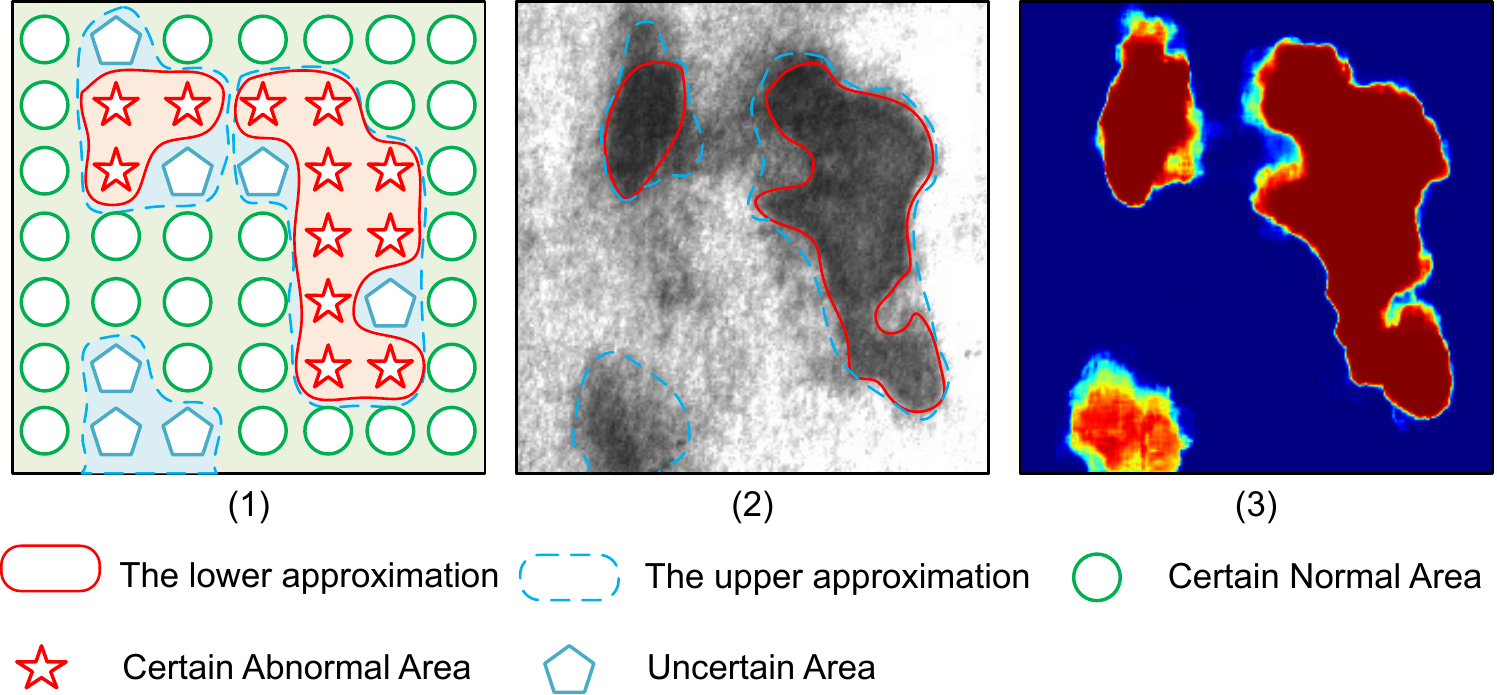}
  \caption{Uncertain model of noisy label based on Rough Set. The suspicious regions are described by lower and upper approximations in (1) and (2). The segmentation probability that is solvable for neural networks is used to characterize uncertain regions in (3).}
  \label{Fig. 2}
  \end{figure} 

As for the first aspect, the existing methods avoid the negative effects by correcting or evaluating noisy labels. However, they have some limitations. GAT \cite{Yi2022Learning}, WSSS \cite{Lu2017Learning}, MV-DAR \cite{Liao2022Learning}, and MTCL \cite{Xu2022Anti} learn segmentation models by introducing additional label information, such as few clean labels, intermediate labeling variable, and multi-labels, which is time-consuming and laborious. And due to the indistinguishability, it is difficult to obtain the suspicious regions labels or clean/true labels. ADL \cite{Wang2021RAR} and Pick-and-Learn \cite{zhu2019pick} proposed label quality evaluation to adjust the loss of training example. However, they evaluate labels at image-level, which is not as fine-grained as pixel-level. In addition, since the noisy labels for defect detection are characterized by inconsistencies rather than falsities, designing an evaluation strategy is challenging.

The key to learn a trustworthy model from noisy labels is how to deal with the suspicious regions. Therefore, based on Rough Set \cite{pawlak1982rough}, we define the suspicious regions as uncertain regions, which are the "third regions" in addition to the normal and abnormal regions. And the uncertain regions are represented by tow precise boundary lines, which are called the lower approximation and the upper approximation, as illustrated in Fig. 2. The lower approximation consists all regions which surely belong to the anomalies, while the upper approximation contains all regions which possibly belong to the anomalies. The uncertain regions of the image consist all pixels that cannot be uniquely classified by employing available features. Specifically, the segmentation probabilities solvable by neural networks are used to characterize uncertain regions, where the pixel value is the probability that the pixel is abnormal, as shown (3) of Fig. 2. Finally, the lower and upper approximation are precise and are without uncertainty and inconsistency. Therefore, inspired by tversky loss \cite{salehi2017tversky}, we redesign the loss function, that the lower approximation is used to calculate the precision penalty and the upper approximation is used to calculate the recall penalty.

With the regard to the second aspect, standard deep learning models for segmentation are not able to capture uncertainty. Bayesian probability theory provides a mathematical tool to reason about model uncertainty. Monte-Carlo dropout \cite{gal2015bayesian, gal2016dropout} is used as approximate Bayesian inference on network weights, approximating the posterior distribution by sampling from a Bernoulli distribution. But there are still some problems. Dropout \cite{kendall2015bayesian, Hiasa2020Automated} is applied after convolution layers to establish Bayesian neural networks in many methods, which randomly masks some pixels. However, a single pixel has no semantic, while normal or abnormal are both context-dependent semantic descriptions. In addition, the existing Dropout based Bayesian neural networks (BNNs) are proprietary designs, which are difficult to generalize to other networks directly. Neural networks in industrial scenarios are often customized, so pluggability is a particularly important prerequisite.

In this paper, both over-labeling and under-labeling exist in uncertain regions with same feature in the dataset. We assume that the distribution of uncertain regions in noisy labels can be obtained by fitting the prior distribution of multiple weights in the Bayesian neural network.  Therefore, based on DropBlock \cite{ghiasi2018dropblock}, we propose PSBM, which drops contiguous regions from layer's feature map instead of dropping out independent random units. And we explore how to apply PSBM block to construct Bayesian neural networks (BNNs) without changing the network structure. Concretely, we discussed how many and where the PSBM should be applied. At last, at training time, the uncertain regions obtained by computing the variance of multiple results of the BNNs are used to correct the labels. When testing, we intersect multiple results of the model to get the lower approximation, and take a union to get the upper approximation. 

Finally, we obtain the semantic segmentation probability, and further obtain the lower approximation, upper approximation and uncertain regions. Then, how to apply these results to obtain indicators that have reference significance for industrial practice?

We propose defect discrimination confidence to measure the uncertainty, with which anomalies can be identified as defects. In the segmentation probability, different geometric dimensions can be obtained by taking different probability. We compare these geometries with the threshold given by national-, industry-, or factory-standard. When the geometric dimensions are smaller than threshold, the confidence is 0\%. When the geometric dimensions are bigger than threshold, the confidence is 100\%. When the threshold is between the geometric dimensions, the confidence is the probability value corresponding to the threshold. In production, confidence level can be used to grade products. And the factory can determine the confidence threshold according to the actual demand so as to obtain the judgment results with good consistency.

In summary, we propose a framework to learn a trustworthy model from noisy labels, which requires no additional labels and changing network structures. The contributions of this paper are as follows.

1) To avoid the negative effects of noisy labels, the representation of suspicious regions and new loss function are proposed. Suspicious regions are represented by precise lower and upper approximation in pixel-level. And the loss function is redesigned by combining precision penalty of lower approximation and recall penalty of upper approximation.

2) We propose pluggable spatially correlated Bayesian module that can produce probabilistic pixel-wise segmentation without changing the network structure and adding any extra labels.

3) Defect discrimination confidence measure of uncertainty is proposed to identify which anomalies are defects, which can be used to grade the appearance of the product.

The remaining part of this paper is organized as follows. The related works are discussed in Section \uppercase\expandafter{\romannumeral2}. Then, in Section \uppercase\expandafter{\romannumeral3}, we describe the procedures of the proposed loss function, PSBM and confidence. The ablation, comparison, application and real-time experiment results are presented in Section \uppercase\expandafter{\romannumeral4}. Finally, we make a conclusion of the paper in Section \uppercase\expandafter{\romannumeral5}. 

\section{Related Works}

Our proposed method is learning trustworthy models from noisy labels, which properly handle the uncertainty and inconsistency introduced by suspicious regions in surface defect detection. Rough Set and Bayesian Neural Networks(BNNs) are applied to the proposed method. This Section briefly reviews the development of Rough Set and BNNs. Then, latest research on learning from noisy labels is discussed.

\subsection{Development of Rough Set and Bayesian Neural Networks}

The basic notions of Rough Sets and approximation spaces are introduced by Pawlak \cite{pawlak1982rough}. Since then, it has often been proved to be an excellent mathematical tool for analyzing vague description of object. The adjective vagueness is concerned with inconsistency and ambiguity. Any subset $X$ of the universe $U$ can be expressed in terms of elementary sets either precisely or approximately. In the latter case, the subset $X$ can be characterized by two ordinary sets, which are called the lower approximations and upper approximations. The lower approximation of $X$ is composed of all elementary sets contained in $X$, while the upper approximation of $X$ consists all the elementary sets that have a non-empty intersection with $X$. The difference between the upper and lower approximation constitutes the boundary region of the Rough Set, whose elements cannot be characterized as belonging to $X$ or not with certainty. In this paper, the suspicious region is also expressed approximately. The lower approximation consists all regions which surely belong to the defects, while the upper approximation contains all regions which possibly belong to the defects.

BNNs \cite{neal2012bayesian} provide probabilistic interpretations of deep learning models by inferring distributions of the model weights that converge to Gaussian processes. However, because of the vast number of parameters, model-ling a distribution over the kernels is a challenging inference and will bring additional computational costs. Yarin \cite{gal2016dropout} show that the use of dropout in neural networks can be interpreted as a Bayesian approximation of Gaussian processes without sacrificing either computational complexity or test accuracy. Meanwhile, Yarin \cite{gal2015bayesian} propose a new practical dropout CNN architecture. Dropout networks' training can be cast as approximate Bernoulli variational inference in Bayesian Neural Networks, while the model can be evaluated by approximating the predictive posterior - referred as Monte Carlo dropout in test time. Alex \cite{kendall2015bayesian} present a deep learning framework for probabilistic pixel-wise semantic segmentation, which is called Bayesian SegNet. In this paper, we propose a surface defect detection network based on BNN as an extending to the previous works.

\subsection{Latest Research on Learning from Noisy Labels Methods}

The noisy labels in real-word datasets are reported to range from 8.0\% to 38.5\% \cite{Song2022Learning}. Learning from noisy labels is becoming an important task and has grained more and more attention. To learn from noisy labels, robust architecture \cite{srivastava2014dropout, Xiao2015Learning, Yao2019Deep, Batch2015Batch} adds a noise adaptation layer at the top of the softmax layer and designs a new dedicated architecture. Robust regularization methods \cite{Jenni2018Deep, Lukasik2020Does} improve the robustness to label noise with widely-used regularization techniques, such as data augmentation, weight decay and dropout. Robust loss functions \cite{zhang2018generalized, wang2017multiclass} are designed to achieve a small risk for unseen clean data even when noisy labels exist in the training data. Sample selection methods \cite{Han2018Co-teaching, Huang_2019_ICCV} propose evaluation strategies to select true-labeled examples from a noisy training dataset. But most of these methods only consider image-level labels instead of pixel-level ones. 

As for pixel-level, GAT \cite{Yi2022Learning} correct the noisy labels using a graph attention network supervised by detected clean labels. WSSS \cite{Lu2017Learning} learned segmentation models from noisy labels by introducing an intermediate labeling variable. ADL \cite{Wang2021RAR} proposed an adaptive denoising learning strategy to avoid influence from the noisy labels. Pick-and-Learn \cite{zhu2019pick} introduce a label quality evaluation strategy to let the network maximally learn from the clean annotations during the training process. However, due to the indistinguishability of the suspicious regions, it is difficult to obtain clean/true labels or to evaluate the noisy labels.

\section{Proposed Approach}

\subsection{System Overview}

The architecture of the proposed approach includes three major components: i) the representation of noisy labels and the redesigned loss function; ii) pluggable spatially correlated Bayesian module (PSBM) and it's application mode; iii) defect discrimination confidence.

1) At first, we will discuss the basic concepts of representations about noisy labels based on Rough Set. And the redesigned loss functions will be introduced.

2) Secondly, the probabilistic model and PSBM will be illustrated. Then, in order to apply the proposed PSBM to various customized models, pluggable application modes will be introduced. Training mechanism and inference of segmentation probability will be described. At last, we take U-net as an example to obtain a U-net based BNNs. 

3) Finally, we will introduce the calculation method of discrimination confidence.

\subsection{The Representation of Noisy Labels and redesigned loss function}

Suspicious regions are indistinguishable when annotating, and noisy labels are inconsistent. Rough set has been proved to be an excellent mathematical tool for the analysis of an inconsistent description of object. Consequently, we represent and correct the noisy labels based on Rough set.

\subsubsection{Representation of noisy labels based on Rough Set}

Consider a simple knowledge representation of noisy labels in which a finite set of regions is described by a finite set of attributes. Formally, it can be defined by an information system $S$ expressed:
\begin{equation}
  \label{E1}
  S=(U,A)
  \end{equation}
where $U$ is a finite nonempty set of all regions in the image, and $A$ is a finite nonempty set of attributes (such as texture, grayscale, etc). 

\textbf{Definition 1. (Indiscernible relation)} Given a subset of attribute set about anomalies $B\subseteq A$, the indiscernible relation $ind(B)$ on the universe $U$ can be defined as follows,
\begin{equation}
  \label{E2}
  ind(B) = \{(x,y)\mid (x,y)\in U^2, \forall_{b\in B} (b(x)=b(y))\}
  \end{equation}

The equivalence relation is an indiscernible relation. And the equivalence class of $x$ is denoted by $[x]_{ind(B)}$, or simply $[x]$.

\textbf{Definition 2. (Upper and lower approximation sets)} For an anomaly region $R\subseteq U$, its lower and upper approximation sets are defined respectively, by
\begin{equation}
  \label{E3}
  \overline{apr}(R)=\{r\in U \mid [r]\cap R=\emptyset \}
  \end{equation}
\begin{equation}
  \label{E4}
  \underline{apr}(R)=\{r\in U \mid [r] \subseteq R\}
  \end{equation}
where $[r]$ denotes the equivalence class of $r$.

\textbf{Definition 3. (The anomaly, boundary and normally regions)} The family of all equivalence classes is known as the quotient set of $U$, and it is denoted by $U/A=\{[r]\mid r\in U\}$. The universe can be divided into three disjoint regions, namely, the anomaly, boundary and normally regions,
\begin{equation}
  \label{E5}
  ANO(R)=\underline{apr}(R)
  \end{equation}
\begin{equation}
  \label{E6}
  NOR(R)=U-\overline{apr}(R)
  \end{equation}
\begin{equation}
  \label{E7}
  BND(R)=\overline{apr}(R)-\underline{apr}(R)
  \end{equation}
If a region $r\in ANO(R)$, then it belongs to anomalies set $R$ certainly. If a region $r\in NOR(R)$, then it doesn't belong to $R$ certainly. If a region $r\in BND(R)$, then it cannot be determined whether the region $r$ belongs to $R$ or not. That is, the suspicious regions are represented by $BND(R)$.

\textbf{Definition 4. (Segmentation probability)} For getting a solvable representation for neural networks, we define the Segmentation probability to characterize the suspicious regions, where the pixel value is the probability that pixel is anomalous.
\begin{equation}
  \label{E8}
  \underline{apr}(R)=\{v\in V \mid v = 1\}
  \end{equation}
\begin{equation}
  \label{E9}
  \overline{apr}(R)=\{v\in V \mid v > 0\}
  \end{equation}
where $v$ is the pixel value, $V$ is a set of all pixels in image.

\subsubsection{Redesigned loss function}

Although there is inconsistency in the suspicious region, the lower and upper approximation are consistent. Therefore, inspired by tversky loss \cite{salehi2017tversky}, we redesign the loss function as follows:
\begin{equation}
  \begin{split}
  \label{E10}
  Loss = 1-\alpha \frac{\underline{apr}(Y)\cap \mathcal{Y}}{\underline{apr}(Y)\cap \mathcal{Y} + \left\lvert \mathcal{Y}-\underline{apr}(Y)\right\rvert }\\
  -\beta \frac{\overline{apr}(Y)\cap \mathcal{Y}}{\overline{apr}(Y)\cap \mathcal{Y} + \left\lvert \overline{apr}(Y) - \mathcal{Y} \right\rvert}
  \end{split}
  \end{equation}
where $Y$ is the label, $\mathcal{Y}$ is the result of neural networks, and $\alpha +\beta =1$. The lower approximation is used to calculate the precision penalty and the upper approximation is used to calculate the recall penalty.

\subsection{The methods and application of pluggable spatially correlated Bayesian module}

\subsubsection{Probabilistic modelling} Given the training inputs $\{x_1,...,x_N\}$ and their corresponding labels $\{y_1,...,y_N\}$, we would like to estimate a function $y=f(x)$. We would put some prior distribution over the space of functions $p(f)$ in the Bayesian approach. And the posterior distribution is looked for overing the space of functions given our dataset: $p(f \mid X, Y)$.

In Bayesian neural network, we are interested in finding the posterior distribution over the convolutional weights
\begin{equation}
  \begin{split}
  \label{E11}
  w=(W_i)_{i=1}^{I}
  \end{split}
  \end{equation}
where $W_i$ is the weight of the $i^{th}$ layer convolutional network. But the distribution $p(w\mid X, Y)$ is not tractable. Therefore, we define an approximation variational distribution $q(w)$ to approximate $p(w)$. Inspired by \cite{gal2015bayesian}, we use Gaussian prior distributions to approximate $q(w)$. Then the Gaussian process can be approximate by Bernoulli distributed random variables with dropout probabilities $b_{i,j}$ and variational parameters of CNN's kernels $K_i$. $b_{i,j}$ is the dropout probability of the $j^{th}$ neuron of the $i^{th}$ layer network. $K_i$ is the convolutional kernel of the $i^{th}$ layer network. Thus the $q(W_i)$ is defined for every layer $i$ as
\begin{equation}
  \begin{split}
  \label{E11}
  &W_i=K_i \cdot diag([b_{i,j}]_{j=1}^J)\\
  &b_{i,j} \sim Bernoulli(p_i) for \ i=1,...,I, j=1,...J
  \end{split}
  \end{equation}
The $diag(\cdot )$ operator maps vectors to diagonal matrices whose diagonals are the elements of the vectors. $p_i$ is a fix Bernoulli distribution probability. In general, we set $p_i=0.5$. The network has a total of $I$ layers, and each layer has $J$ neurons.

In network optimization, the distribution over the network's weights is obtained by minimizing the Kullback-Leiber (KL) divergence between two distributions:
\begin{equation}
  \label{E12}
  KL(q(w) \| p(w \mid X, Y))
  \end{equation}
Minimizing the KL divergence is equivalent to maximizing the \textit{log evidence lower bound}:
\begin{equation}
  \label{E13}
  L_{\uppercase\expandafter{\romannumeral4}}:=\sum_{i=1}^{N}E(y_i,\hat{f}(x_i, \hat{w_i}))-KL(q(w) \| p (w))
  \end{equation}
where $E(\cdot )$ is a softmax likelihood loss function, and $\hat{w_i}\sim q(w)$. According to \cite{gal2016dropout}, we use $L_2$ regularization to weight:
\begin{equation}
  \label{E14}
  L_{dropout}:=\sum_{i=1}^{N}E(y_i,\hat{f}(x_i, \hat{w_i}))+ \sum_{i = 1}^{I}(\left\lVert W_i\right\rVert _2^2 + \left\lVert b_i\right\rVert _2^2 )    
  \end{equation}
where $\left\lVert \cdot \right\rVert _2^2$ is the square of $L_2$ norm.

In networks inference, we approximate the integral with Monte Carlo integrations:
\begin{equation}
  \label{E15}
  p(y^* | x^*, X, Y) \approx \frac{1}{T}\sum_{t = 1}^{T}\hat{f}(x^*, \hat{w_t})     
  \end{equation}
where $x^*$ and $y^*$ are the input and output in test set, and $\hat{w_t} \sim q(w)$. $T$ is a hyperparameter used to balance the accuracy of calculation results and computational overhead.
\subsubsection{The PSBM and it's application modes}

\begin{algorithm}[t]
  \caption{PSBM}
  \label{A1}
  \begin{algorithmic}[1]
    \Require
      Feature map: $A_{input}$ of size $(C,H,W)$;
      Convolution kernel: $K$ of size $(L,L)$;
      Probability: $p=0.5$.
    \Ensure
      Output feature map: $A_{output}$.
    \State Bernoulli distribution probability of PSBM $\gamma$ :\\
    \centerline{$\gamma  = ((1-p)/L^2)\cdot (W^2/(W-L+1)^2)$}
    Dropout probabilities $b_i$:\\
    \centerline{$b_j \sim Bernoulli(\gamma )$ for $j=1,...,L^2$} 
    Randomly sample Mask $M$: \\
    \centerline{$M=diag([b_j]_{j=1}^{L^2})$} 
    Get Block mask $M_{Block}$ by max pooling, pooling size is $(L,L)$, stride is $(1,1)$, and padding is $(L/2, L/2)$: \\
    \centerline{$M_{Block}=1-max\_pool(M)$} 
    Apply the Block Mask: \\
    \centerline{$A_{output}=A_{input}\times M_{Block}$} 
    Normalize the features:\\
    \centerline{$A_{output}=A_{output}\times cout(M_{Block})/cout\_ones(M_{Block})$} 
    return $A_{output}$
  \end{algorithmic}
\end{algorithm}

Based on DorpBlock \cite{ghiasi2018dropblock}, we design the PSBM. Pseudocode of PSBM is illustrated in algorithm \ref{A1}. In order to improve computing efficiency, the entire process of PSBM preforms tensor calculations in the GPU. Specifically, the Mask $M$ is obtained through the function $torch.Bernoulli$. And the Block Mask $M_{Block}$ is obtained through maximum pooling. Furthermore, given that both the probability $p$ and feature $A_{output}$ are normalized, the latter term of (\ref{E14}) can be regarded as $0$.

We explore how to apply PSBM to construct BNNs without changing the networks structure. Concretely, we discuss the following problems:

(a) How many PSBM should be applied. In practice, Dropout randomly drop units from the neural networks, preventing overfitting but weakening the network's ability to learn. And over-powerful regularization will make the networks learn slowly. Therefore, we follow two principles. Firstly, PSBM should not be applied at every layer of the network as a regularization method. Secondly, PSBM should not co-occur with other regularization methods in the same layer network.

(b) Where PSBM should be applied. As we know, existing semantic segmentation models basically consist of encoders and upsampling layers. Therefore, the layers of networks are divided into low-level and high-level layers. In general, low-level network extracts low-level features, such as edges and corners, high-level network extracts high-level features, such as shape and contextual relationships. As we know, low-level features are consistent and shared across the distribution of models, while high-level features are masked by PSBM may be more effective. In summary, PSBM should be applied at the high level of the encoders.

\subsubsection{Training mechanism}
When training, inputs of the training set are represented by $\{x_1,...,x_N\}$, and $\{y_1,...,y_N\}$  are their annotation. $\{\hat{y}_i^1,...,\hat{y}_i^{\mathcal{T}}\}$ are the segmentations of $i^{th}$ samples in training set, which are calculated $\mathcal{T}$ times by BNNs:
\begin{equation}
  \label{E21}
  \hat{y}_i^t = \hat{f}(x_i, \hat{w}_t)
  \end{equation}
Initially, the mean $\mu _i$ and variance $\sigma_i$ of the segmentations are
\begin{equation}
  \label{E22}
  \mu_i=\frac{1}{\mathcal{T}}\sum_{t = 1}^{\mathcal{T}}(\hat{y}_i^t)   
  \end{equation}
\begin{equation}
  \label{E23}
  \sigma _i=\frac{1}{\mathcal{T}}\sum_{t = 1}^{\mathcal{T}}(\hat{y}_i^t - \mu _i)   
  \end{equation}
And the variance $\sigma_i$ are normalized on pixel-level
\begin{equation}
  \label{E24}
  \hat{\sigma_i}=(\sigma_i-min(\sigma_i))/(max(\sigma_i)-min(\sigma_i))
  \end{equation}
In general, the variance represents the uncertain part of the label, which is the suspicious region. Therefore, we correct the labels with normalized variances:
\begin{equation}
  \label{E25}
  \underline{apr}(y_i) = y_i- y_i\times \hat{\sigma_i}
  \end{equation}
\begin{equation}
  \label{E26}
  \overline{apr}(y_i) = y_i + \hat{\sigma_i}
  \end{equation}
Finally, according to equation (\ref{E10}) and (\ref{E14}), the loss function is
\begin{equation}
  \begin{split}
  \label{E27}
  Loss = \frac{1}{N}\sum_{i=1}^{N}(1-\alpha \frac{\underline{apr}(y_i)\times \hat{y}_i^t}{\underline{apr}(y_i)\times \hat{y}_i^t + (1-\underline{apr}(y_i))\times \hat{y}_i^t}  \\
  - \beta \frac{\overline{apr}(y_i)\times \hat{y}_i^t}{\overline{apr}(y_i)\times \hat{y}_i^t +\overline{apr}(y_i) \times (1-\hat{y}_i^t) } )
  \end{split}
  \end{equation}
where $\alpha +\beta =1$, $\hat{y}_i^t$ is randomly selected from $\{\hat{y}_i^1,...,\hat{y}_i^{\mathcal{T}}\}$.

\subsubsection{Inference for segmentation probability}

We approximate the segmentation probability with Monte Carlo integrations. Input and output of the testing set is represented by $x^*$ and $y_1^*$. On the basis of equation (15), the probability is calculated as follows:
\begin{equation}
  \label{E28}
  p(y^*) \approx \frac{1}{T}\sum_{t = 1}^{T}\hat{f}(x^*, \hat{w}_t)   
  \end{equation}

\subsubsection{U-net based Bayesian neural networks}

\begin{figure*}[!t]
  \centering
  \includegraphics[width=6.8in]{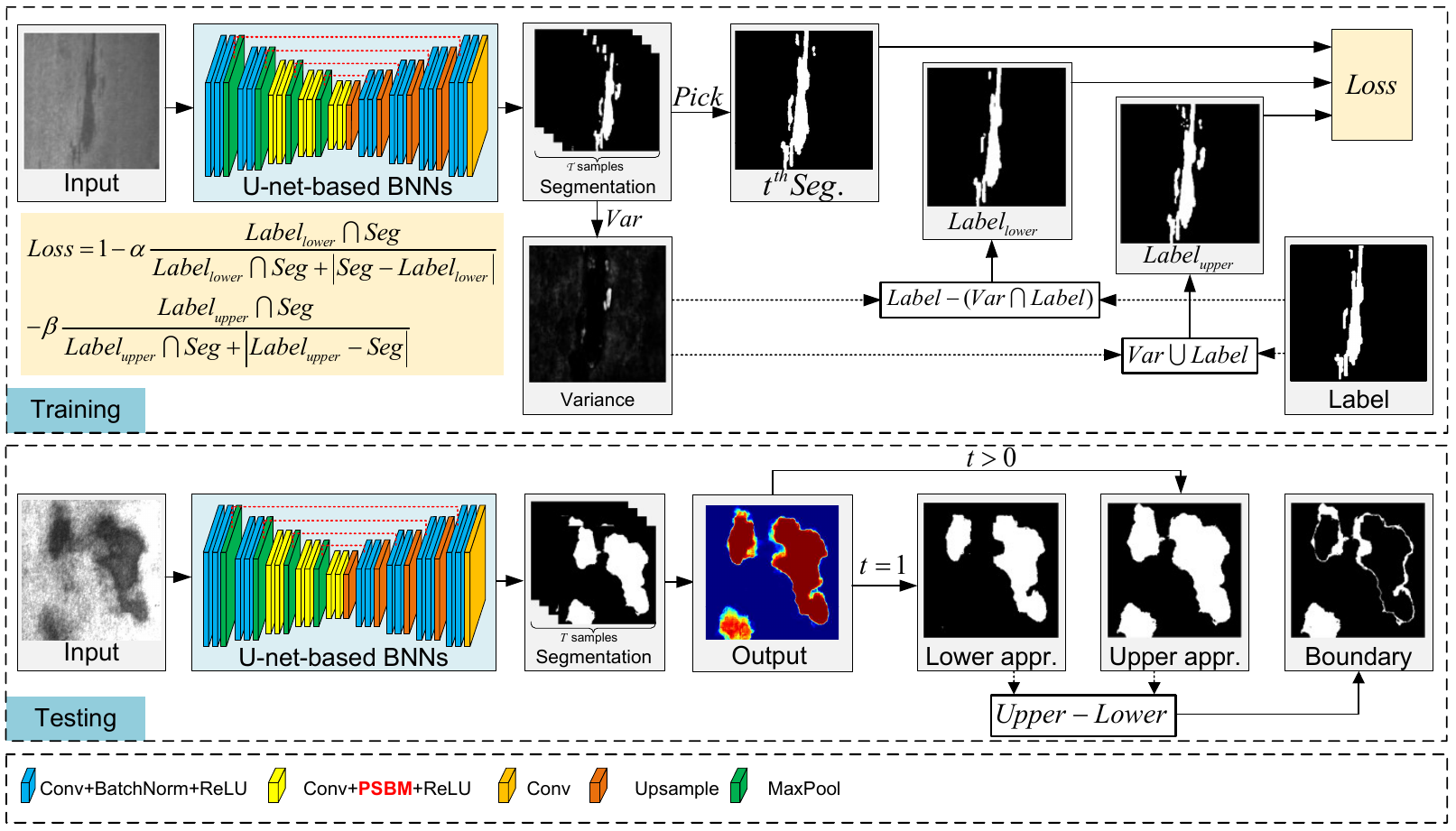}
  \caption{An overview of optimization and inference of U-Net-based BNNs transformed by pluggable Bayesian modules.}
  \label{Fig. 4}
  \end{figure*}

To show the optimization and inference of BNNs more clearly, we design the overall structure of the network (as illustrated in Fig. \ref{Fig. 4}.) based on U-net \cite{ronneberger2015u}. According to the applied method in the subsection 2), we replace BatchNorm in the last three layers of the encoder with PSBM. 

When training, input is calculated multiple times by U-net-based BNNs to obtain N samples Segmentations (Seg). The noisy labels are corrected by the variance (Var) of multiple Segs, resulting in lower approximation ($Label_{lower}$) and upper approximation ($Label_{Upper}$). Then $i^{th}$ Seg, ($Label_{lower}$) and ($Label_{Upper}$) are used to calculate the loss function. 

When testing, we approximate the segmentation probability (Output) with Monte Carlo integrations, as shown in equation (\ref{E23}). Then, according to the Output, the lower, upper approximation, and boundary are obtained.

\subsection{The calculation method of discrimination confidence}
\begin{figure}[!t]
\centering
\includegraphics[width=3.4in]{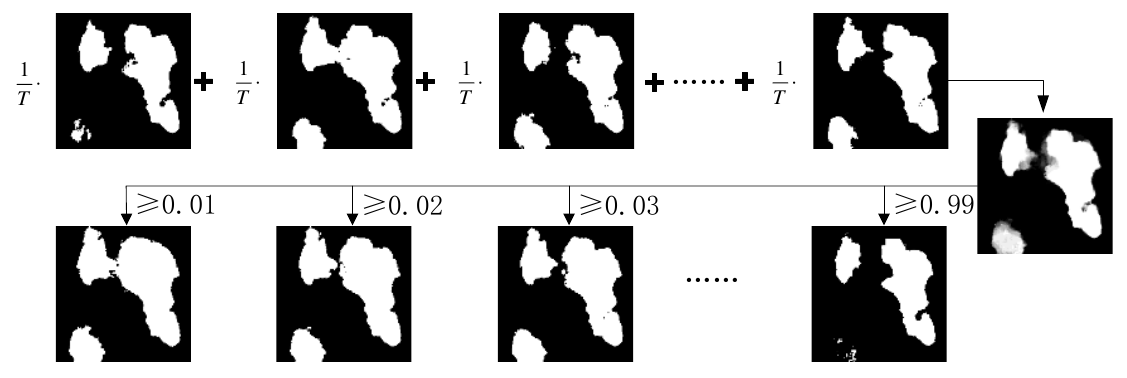}
\caption{Confidence evaluation method.}
\label{Fig. 6}
\end{figure}
Now, we get the probability of each pixel in the image. Then, it is crucial to calculate the confidence of networks discriminations based on the probability. As we all know, the semantic segmentation results cannot be used directly to distinguish whether the sample is NG or not. The geometric dimensions of the defective region, such as length, diameter, etc., need to be counted. Finally, according to the threshold given by the national-, industry-, or factory-standard, it is determined whether it is defective or not. Therefore, we define the discriminant confidence based on the relationship between the threshold and the probability.

As illustrated in Fig. 6, we take different probabilities $\lambda \in [0,1]$. According to equation (\ref{E8}) and (\ref{E9}), when $\lambda=1$, the lower approximation is obtained, while $\lambda>0$, the upper approximation is given. And $v$ is the value of each pixel of the segmentation. We define that the geometric dimensions ($GD$) of the defective region are calculated by
\begin{equation}
  \label{E30}
  g(\lambda)=\int_{v=\lambda}^{v=1} GD(v) \,dv
  \end{equation}
And the threshold given by the national-, industry-, or factory-standard is represented by $\Lambda $. The confidence is calculated as follows:
\begin{equation}
  \label{E31}
  C_{x^*}(\Lambda)=\left\{
  \begin{aligned}
    &0\%, &&g(0^+)<\Lambda \\
    &g^{-1}(\Lambda), &&g(1)<\Lambda <g(0^+) \\
    &100\%, &&g(1)>\Lambda 
  \end{aligned}\right.
  \end{equation}
where $g^{-1}(\Lambda)$ represent the inverse function of $g(\Lambda)$.

\section{Experiments and Results}
\subsection{Implementation Details}
\subsubsection{Parameters setting}
The base learning rate is 0.003 with a decay of 0.0001, and the mini-batch size is 4. And the standard probability of dropping a connection is set as 50
\subsubsection{Computation Platform}
The proposed method is implemented on the PyCharm with the open source toolbox Pytorch. In addition, we train the model on the high-performance server, NVIDIA Tesla A100 GPU (with 40G memory) on CentOS 8 Linux. 

\subsection{Datasets}

\begin{table}
  \begin{center}
  \caption{Number of three datasets.}
  \label{tab1}
  \begin{tabular}{| c | c | c |}
  \hline
  Dataset & Train & Test\\
  \hline
  NEU-seg& 3630 & 840\\
  \hline
  MCSD-seg& 532 & 134\\ 
  \hline
  LC-seg& 502 & 126\\
  \hline 
  \end{tabular}
  \end{center}
  \end{table}

In this paper, three datasets are selected to support and evaluate the applicability and generality of the proposed method. They are one benchmark dataset, NEU-seg \cite{song2013noise} and two datasets obtained from the reality industrial line, including MCSD-seg \cite{Niu2022Positive} for motor commutator, LC-seg for Light chip. All images are resized to $256 \times 256$, and the training and test-ing sets are randomly divided by 8:2, as illustrated in table \uppercase\expandafter{\romannumeral1}. 

\subsection{Evaluation Metrics}
The noise occurs in almost every label. It is difficult to obtain clean/true labels. Therefore, 
to demonstrate the ability of the network to solve the lower and upper approximation, we use the lower approximation to calculate accuracy and the upper approximation to calculate recall rate. And to evaluate the segmentation ability, the intersection-over-union (IoU) is used to evaluate the performance compared with other segmentation methods. 

\subsection{Ablation Experiment}

In this paper, we mainly made the following improvements at the network level: 1) proposed the PSBM based on DropBlock; 2) obtained the upper and lower approximations by Bayesian inference; 3) redesigned the loss function based on Rough set. Therefore, we design four sets of ablation experiment based on U-net \cite{ronneberger2015u} on NEU-seg: 1) the original U-net is used as the basic control group; 2) the network obtained by modifying U-net based on PSBM is used as the second control group; 3) on the basis of the second control group, Bayesian inference is added as the third control group; 4) our method, as the fourth group, improves U-net based on PSBM, Bayesian inference and rough set based loss function.

The results of ablation experiments are shown in table \uppercase\expandafter{\romannumeral2} and figure 5. PSBM improves the IoU value of U-net, but there is little improvement in recall rate and precision rate. This means that although Dropout can improve the network's ability to fit data, it does not work well for noisy label in surface defect detection. 
The posterior distribution obtained by Bayesian inference can significantly improve the recall rate and precision rate. The false detection regions are indicated with lower probabilities, while the missing detection regions are detected. This shows that Bayesian inference can effectively capture uncertainty in labels.
The loss function based on Rough set further improves the recall rate and precision rate. And the anomalies are also more clearly outlined and probabilistically more accurate.

\begin{table}
  \begin{center}
  \caption{Result of ablation example}
  \label{tab1}
  \begin{tabular}{| c | c  c  c |}
  \hline
   & \multicolumn{3}{c}{NEU-seg} \vline \\
   & Recall & Precision & IoU\\
  \hline
  U-net& 0.8845 & 0.8455 & 0.7581\\
  \hline
  U-net+PSBM & 0.8802 & 0.8563 & 0.7644\\ 
  \hline
  U-net+PSBM+Bayes& 0.9113 & 0.9175 &0.7643 \\
  \hline
  U-net+PSBM+Bayes+Loss functions& 0.9350 & 0.9390 & 0.7670\\
  \hline
  \end{tabular}
  \end{center}
  \end{table}

\begin{figure}[!t]
  \centering
  \includegraphics[width=3.4in]{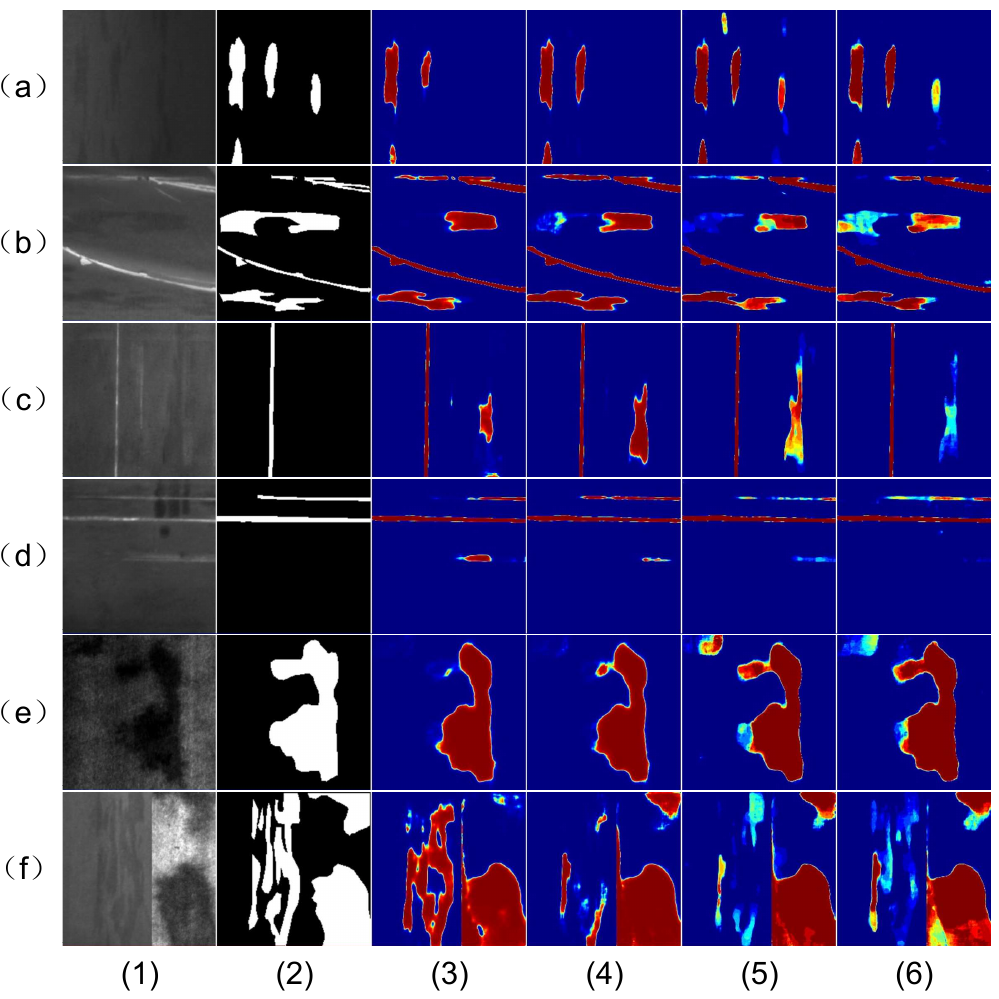}
  \caption{Results of Ablation experiment. Column (1) and (2) are the images and labels. Column (3), (4), (5) and (6) are the results of original U-net, improved U-net based on PSBM, improved U-net based on PSBM and Bayesian inference, and ours methods (improved by PSBM, Bayesian inference and Rough set based loss function)}
  \label{Fig. 6}
  \end{figure}  

\subsection{Comparative Experiment}

In this paper, we focus on anomaly segmentation for surface defect detection. Methods \cite{srivastava2014dropout, Xiao2015Learning, Yao2019Deep, Batch2015Batch, Jenni2018Deep, Lukasik2020Does,zhang2018generalized, wang2017multiclass, Han2018Co-teaching, Huang_2019_ICCV} for learning classification models from noisy labels cannot be compared with. As for pixel-level, the methods that learn segmentation models by introducing additional label information are difficult to reproduce in defect detection datasets. Therefore, we compare our methods with ADL \cite{Wang2021RAR} and Pick-and-Learn \cite{zhu2019pick}. In addition, since our method is implemented based on Dropout, we added Dropout noise model \cite{srivastava2014dropout} as a comparison method. In order to control the variables, all the methods above are obtained based on U-net modification. 

As illustrated in table \uppercase\expandafter{\romannumeral2} and figure 6. In quantitative analysis (table \uppercase\expandafter{\romannumeral2}), existing methods have little improvement in recall rate, prediction rate and IoU. This phenomenon proves that evaluating noisy labels at the image-level is inaccurate, and evaluating the inconsistency of noisy labels is very challenging. Combined with the results of qualitative analysis (figure 6), our method can achieve high recall and precision rates, avoiding false and missing defections of abnormal regions. It is efficient to explore and exploit accuracy and consistent elements in noisy label for training models, rather than trying to evaluate and correct noisy labels.

\begin{table*}
  \begin{center}
  \caption{Results of Comparative Experiments}
  \label{tab1}
  \begin{tabular}{| c | c c c | c c c | c c c |}
  \hline
   & \multicolumn{3}{c}{NEU-seg} \vline & \multicolumn{3}{c}{MCSD-seg} \vline & \multicolumn{3}{c}{LC-seg} \vline \\
   & Recall & Precision & IoU & Recall & Precision & IoU & Recall & Precision & IoU \\
  \hline
  U-net\cite{ronneberger2015u}& 0.8845 & 0.8455 & 0.7581 & 0.8234 & 0.8590 & 0.7245 & 0.8572 & 0.8894 & 0.7688\\
  \hline
  Dropout noise model\cite{srivastava2014dropout}& 0.8755 & 0.8535 & 0.7587 & 0.8131 & 0.8760 & 0.7247 & 0.8051 & 0.9148 & 0.7444\\
  \hline
  ADL\cite{Wang2021RAR} & 0.8730 & 0.8556 & 0.7583 & 0.8203 & 0.8663 & 0.7214 & 0.8401 & 0.8888 & 0.7505\\ 
  \hline
  Pick-and-learn\cite{zhu2019pick}& 0.8639 & 0.8633 & 0.7569 & 0.8427 & 0.8464 & 0.7314 & 0.8401 & 0.8643 & 0.7407 \\
  \hline
  Ours& 0.9350 & 0.9390 & 0.7670 & 0.8881 & 0.8978 & 0.7386 & 0.8978 & 0.9401 & 0.7825\\
  \hline
  \end{tabular}
  \end{center}
  \end{table*}

\begin{figure}[!t]
  \centering
  \includegraphics[width=3.4in]{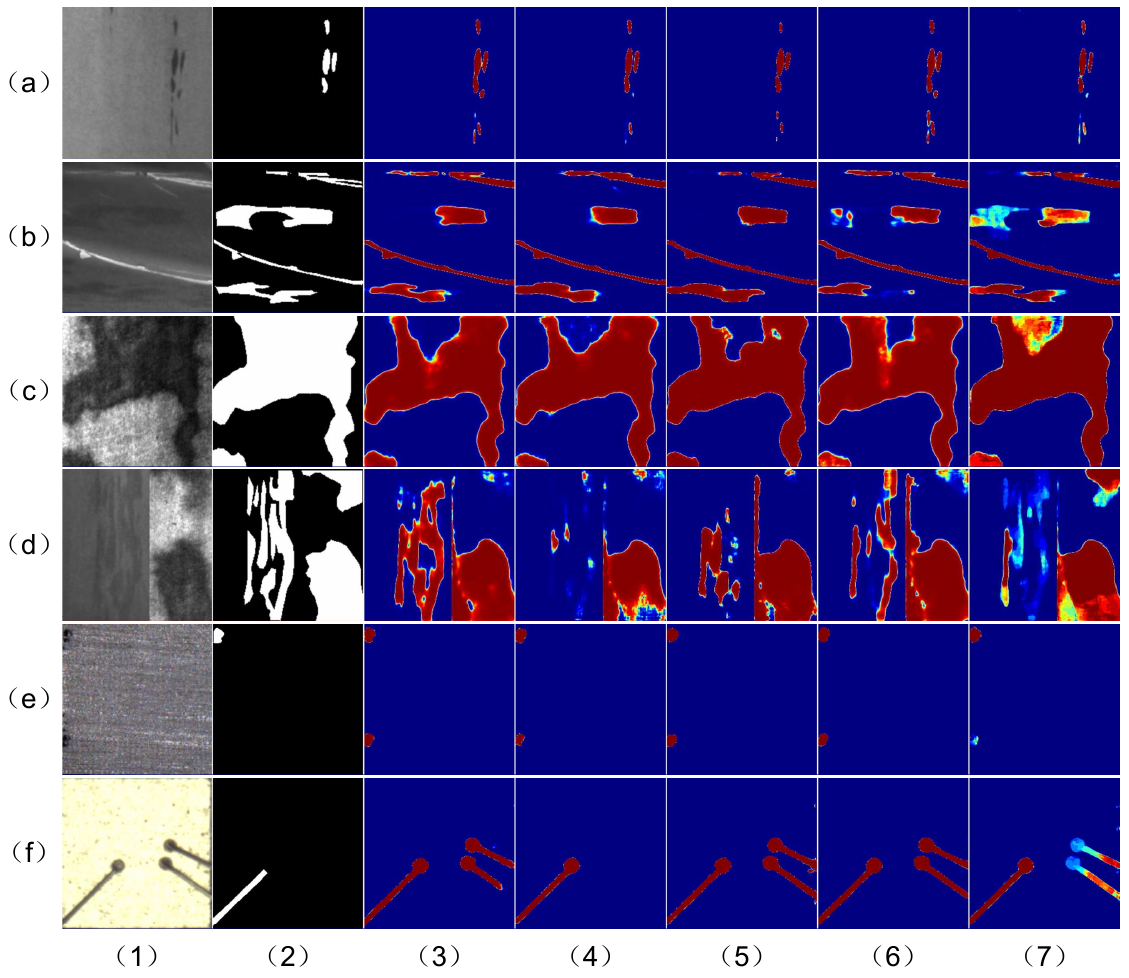}
  \caption{Results of comparative experiment. Column (1) and (2) are the images and labels. Column (3), (4), (5), (6) and (7) are the results of original Unet, Dropout noise model\cite{srivastava2014dropout}, ADL\cite{Wang2021RAR}, Pick-and-learn\cite{zhu2019pick}, and Ours methods.}
  \label{Fig. 6}
  \end{figure}

\subsection{Application Experiments}

To demonstrate the robustness of pluggability, we transform the classic semantic segmentation model (U-net \cite{ronneberger2015u}, FCN \cite{long2015fully}, SegNet \cite{badrinarayanan2017segnet}, and DeepLabeV3+ \cite{chen2017deeplab}) and classic surface defect detection model (PGA-Net \cite{Dong2020PGA}) into BNNs. To verify the robustness, we conduct experiments on three datasets (NEU-seg, MCSD-seg and LC-seg).

The result as shown in table \uppercase\expandafter{\romannumeral4} and figure 7. Compared with the original models, the modified models improved recall rate, precision rate, and IoU by an average of 3.70\%, 3.72\%, and 1.44\% respectively. It is proved that our method greatly improves the accuracy and recall rate, while enhancing segmentation capabilities. Specifically, there are mainly two improvements: 1) false detection and missing detection are improved. As shown in figure 7, the abnormal regions of false detection and missing detection are represented by probability. And the recall rate of upper approximation and the precision rate of lower approximation significantly increased. 2) the pluggability of our methods is robust. Our method has been applied to five classical models and validated on three datasets, which demonstrates that our method can be used for extensively customized networks in surface defect detection.

\begin{table*}
  \begin{center}
  \caption{Results of Application Experiments}
  \label{tab1}
  \begin{tabular}{|c| c  c  c | c c c | c c c |}
  \hline
   & \multicolumn{3}{c}{NEU-seg} \vline & \multicolumn{3}{c}{MCSD-seg} \vline & \multicolumn{3}{c}{LC-seg} \vline \\
   & Recall & Precision & IoU & Recall & Precision & IoU & Recall & Precision & IoU \\
  \hline
  U-net(2015)& 0.8845 & 0.8455 & 0.7581 & 0.8234 & 0.8590 & 0.7245 & 0.8572 & 0.8894 & 0.7688\\
  U-net Based BNNs & 0.9350 & 0.9390 & 0.7670 & 0.8881 & 0.8978 & 0.7386 & 0.8978 & 0.9401 & 0.7825\\ 
  \hline
  PGA-Net(2020)& 0.8664 & 0.8710 & 0.7675 & 0.8526 & 0.8580 & 0.7381 & 0.9143 & 0.8458 & 0.7762 \\
  PGA-Net Based BNNs& 0.8960 & 0.9013 & 0.7677 & 0.9019 & 0.8939 & 0.7514 & 0.9330 & 0.8524 & 0.7929\\
  \hline
  FCN(2015)& 0.8623  & 0.8567 & 0.7514 & 0.8582 & 0.8479 & 0.7393 & 0.8484 & 0.8761 & 0.7676\\
  FCN Based BNNs& 0.8936 & 0.8973 & 0.7473 & 0.8744 & 0.9169 & 0.7465 & 0.8587 & 0.9043 & 0.7804\\
  \hline
  SegNet(2015)& 0.8353 & 0.8684 & 0.7395 & 0.7744 & 0.8306 & 0.6628 & 0.8336 & 0.8807 & 0.7021\\
  SegNet Based BNNs& 0.9107 & 0.8945 & 0.7532 & 0.8552 & 0.8944 & 0.6858 & 0.8384 & 0.9013 & 0.7491\\
  \hline
  DeepLabeV3+(2018)& 0.8791 & 0.8545 & 0.7651 & 0.8253 & 0.8662 & 0.7391 & 0.8579 & 0.8948 & 0.7300\\
  DeepLabeV3 Based BNNs& 0.9021 & 0.8971 & 0.7696 & 0.8642 & 0.8767 & 0.7358 & 0.8781 & 0.8955 & 0.7781\\
  \hline
  \end{tabular}
  \end{center}
  \end{table*}

\begin{figure*}[!t]
  \centering
  \includegraphics[width=7.0in]{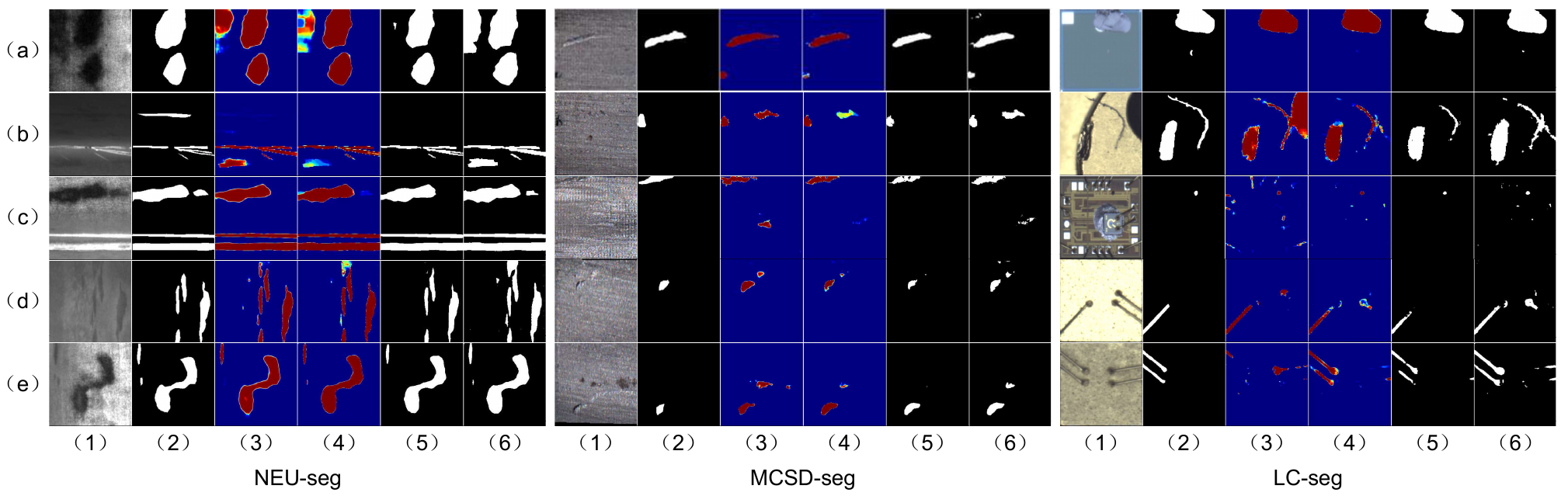}
  \caption{Results of Application examples. Column (a), (b), (c), (d), and (e) are the results of U-net based BNNs, PGA-Net Based BNNs, FCN Based BNNs, SegNet Based BNNs and DeepLabeV3 Based BNNs. Row (1), (2), (3), (4), (5), and (6) are the images, labels, results of original models, segmentation probabilities of BNNs, lower approximations and upper approximations.}
  \label{Fig. 7}
  \end{figure*}

\subsection{PSBM Application Modes Experiments}

In order to verify the application mode of PSBM, we designed location and quantity experiments based on U-net. We design seven groups of comparative experiments: 1) original U-net; 2) PSBM is applied to each layer of encoder; 3) PSBM is applied to each layer of decoder; 4) PSBM is applied to center of U-net; 5) PSBM is applied to the last layer of U-net (Classifier layer); 6) PSBM is applied to each layer of U-net including center, encoder and decoder; 7) PSBM is applied to the center and last two layers of encoder.

The results as shown in table \uppercase\expandafter{\romannumeral5}. Applying PSBM to the encoder and the center layer is helpful for improving representation ability of networks. Recall rate, precision rate and IoU perform best when we place the PSBM in the Center-encoder (last two layers of the encoder and in the center layer). 

\begin{table}
  \begin{center}
  \caption{Results of PSBM application Experiments}
  \label{tab1}
  \begin{tabular}{|c| c  c  c |}
  \hline
   & Recall & Precision & IoU \\
  \hline
  U-net & 0.8845 & 0.8455 & 0.7581 \\
  \hline
  Encoder & 0.9217 & 0.9248 & 0.7631 \\ 
  \hline
  Decoder & 0.9355 & 0.9171 & 0.7562 \\
  \hline
  Center & 0.9037 & 0.9044 & 0.7609 \\
  \hline
  Classifier & 0.6801 & 0.6725 & 0.6375 \\
  \hline
  Center-encoder-decoder & 0.9384 & 0.9179 & 0.7546 \\
  \hline
  Center-encoder & 0.9350 & 0.9390 & 0.7670\\
  \hline
  \end{tabular}
  \end{center}
  \end{table}

\subsection{Real-time Analysis}

\begin{table}
  \begin{center}
  \caption{Result of ablation example}
  \label{tab1}
  \begin{tabular}{| c | c  c | c c |}
  \hline
   & \multicolumn{2}{c}{Original network} \vline & \multicolumn{2}{c}{BNNs} \vline\\
   & Parameters & Times & Parameters & Times\\
   & (MB) & (ms) & (MB) & (ms)\\
  \hline
  U-net& 29.96 & 8.02 & 29.96 & 7.95\\
  \hline
  SegNet & 112.32 & 10.24 & 112.32 & 10.12\\ 
  \hline
  FCN & 85.27 & 9.59 & 85.27 & 9.26 \\
  \hline
  DeepLabeV3+ & 226.37 & 24.87 & 226.37 & 27.88\\
  \hline
  PGA-net & 198.36 & 19.07 & 198.36 & 18.69\\
  \hline
  \end{tabular}
  \end{center}
  \end{table}

The application of automated optical inspection in the production line requires a high level of real-time performance, which means that the models need to be both lightweight and fast. To simulate the case of factory computer, the evaluation of BNN-SDD was conducted on a typical personal computer configuration, NVIDIA GeForce GTX 1070 GPU (with 8G memory), to ensure that the proposed method can be realistically deployed in an industrial setting.
We count the continuous test time of 180 images on the NEU-seg dataset and take the average of the single image for comparison. Moreover, we simulate the actual operation of the production line, taking an image and inferring one.  Although BNNs require 16 computations, we do it in parallel, and get outstanding performance. As is shown in table \uppercase\expandafter{\romannumeral6}, the inference is even faster because PSBM randomly drops some parameters when testing. 
  
\subsection{Application in the production line of motor commutator}

\begin{table}
  \begin{center}
  \caption{Thresholds given by factory-standard in MCSD-seg }
  \label{tab6}
  \begin{tabular}{| c | c  c |}
  \hline
  Dataset & Length (mm) & Width(mm)\\
  \hline
  Tin color & 1.5 & 1.5 \\
  \hline
  Scratches & 2.0 & 2.0\\ 
  \hline
  Indentations & 0.80 & 0.30 \\
  \hline
  Smudge & 0.30 & 0.14\\
  \hline
  \end{tabular}
  \end{center}
  \end{table}

\begin{figure}[!t]
  \centering
  \includegraphics[width=3.0in]{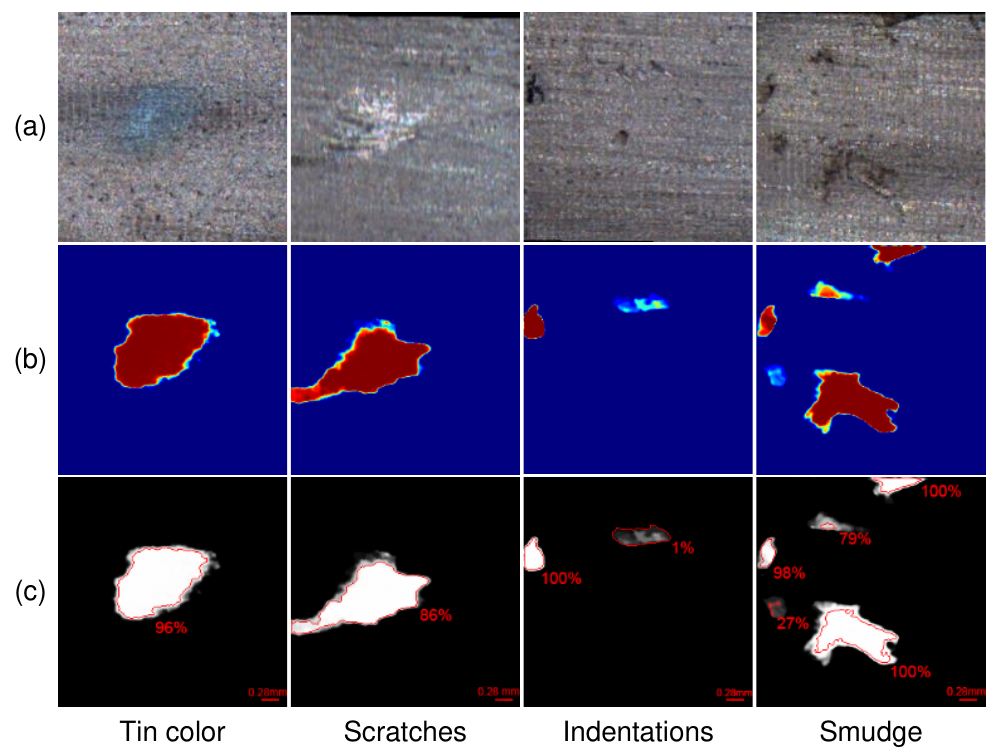}
  \caption{Results of Application in the production line of motor commutator. Row (a), (b), and (c) are the images, segmentation probabilities, and confidences.}
  \label{Fig. 8}
  \end{figure}
  
The defect detection of motor commutators is a typical mental surface defection detection scene. In MCSD-seg used in this paper, there are four types of defects, containing tin color, scratches, indentations and dirt. According to the industry-standard, the thresholds for determining defects are shown in Table \uppercase\expandafter{\romannumeral7}. The size of the image is $256\times 256$, and the pixel equivalent is $0.014mm/pix$.

Firstly, based on U-Net-based BNNs, we complete model optimization and inference, obtaining the segmentation probability of MCSD-seg, as shown in row (b) of Fig. 8. Then, according to formula (24) and (25), the confidence of each connected domain in the probability is obtained, as shown in row (c) of Fig. 8. The red contour lines represent the corresponding confidence anomaly regions whose geometric dimensions reach the threshold and to be judged as defective. The confidence is the probability that the connected domain is judged as a defect. It can be seen that confidence is meaningful for both reliable judgment and product classification.

\section{Conclusion}

We propose a framework to learn a trustworthy model from noisy labels, which requires no additional labels and changing network structures. The noisy labels in surface defect detection are mainly manifested as inconsistencies. Therefore, instead of evaluating and correcting noisy labels, we try to find the precise and consistent elements in noisy labels. Experiments demonstrate that our method is effective for noisy labels in surface defect detection. 

The Rough set based methods for learning from noisy labels can also be applied to medical images. Furthermore, Roughness can be used as a training cost to relabel images to get consist and clean labels. The customized networks can be modified by PSBM to get BNNs, which can not only be used to capture uncertainty, but also to find hard-to-learn features. The discrimination confidence can measure the uncertainty, with which anomalies can be identified as defects. The confidence means that understanding what a model does NOT know.

In the future, we will further explore the factors influencing the noisy label distribution and reduce the requirements for label quality. Particularly, the impacts of label inconsistency on the model convergence rate, and the effect of different imaging channels for inconsistent labels will be further studied.

\bibliographystyle{IEEEtran}
\bibliography{IEEEabrv, myrefs}

\newpage

\section{Biography Section}
\vspace{11pt}

\vspace{-33pt}
\begin{IEEEbiography}[{\includegraphics[width=1in,height=1.25in,clip,keepaspectratio]{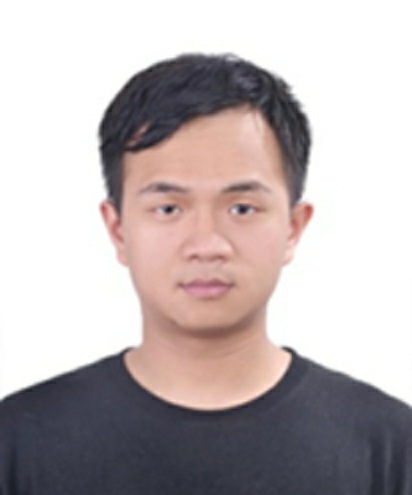}}]{Tongzhi Niu}
received the B.E. degree in mechanical design, manufacturing, and automation from the Wuhan University of Technology, Wuhan, China, in 2018. He is currently working toward the Ph.D. degree with the State Key Laboratory of Digital Manufacturing Equipment and Technology, Huazhong University of Science and Technology, Wuhan. His current research interests include intelligent manufacturing, defects detection, image processing, and deep learning.
\end{IEEEbiography}

\vspace{-33pt}
\begin{IEEEbiography}[{\includegraphics[width=1in,height=1.25in,clip,keepaspectratio]{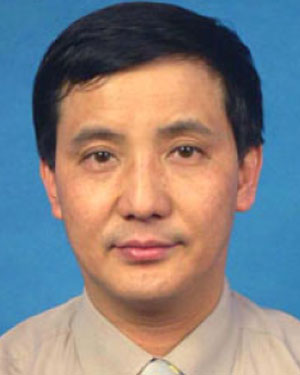}}]{Bin Li}
received the B.E., M.E., and Ph.D. degrees from the Huazhong University of Science and Technology, Wuhan, China, in 1982, 1989, and 2006, respectively, all in mechanical engineering. He is currently a professor with the School of Mechanical Science and Engineering, Huazhong University of Science and Technology. His current research interests include intelligent manufacturing and computer numerical control machine tools.
\end{IEEEbiography}

\vspace{11pt}
\vspace{-33pt}
\begin{IEEEbiography}[{\includegraphics[width=1in,height=1.25in,clip,keepaspectratio]{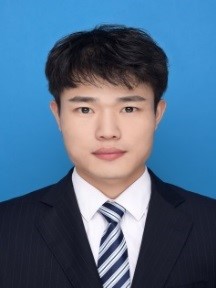}}]{Kai Li}
received the bachelor's degree from the School of Mechanical and Electrical Engineering, Wuhan University of Technology, Wuhan, China. He received the Ph.D. degree in the School of Mechanical Science and Engineering, Huazhong University of Science and Technology, Wuhan. He is currently an  assistant professor with the School of Mechanical and Electrical Engineering, Central South University.
His current research interests include intelligent manufacturing, transfer learning, signal processing, and milling stability.
\end{IEEEbiography}

\vspace{11pt}
\vspace{-33pt}
\begin{IEEEbiography}[{\includegraphics[width=1in,height=1.25in,clip,keepaspectratio]{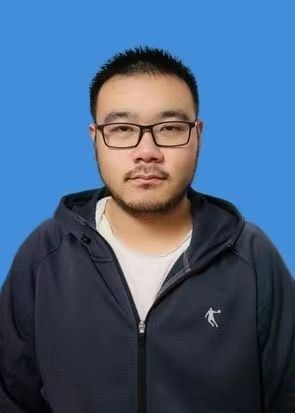}}]{Yufeng Lin}
received the B.E. degree in mechanical design, manufacturing, and automation from the Wuhan University of Technology, Wuhan, China, in 2020. He is currently working toward the M.E. degree with the State Key Laboratory of Digital Manufacturing Equipment and Technology, Huazhong University of Science and Technology, Wuhan. His current research interests include intelligent manufacturing, defects detection, image processing, and deep learning.
\end{IEEEbiography}

\vspace{11pt}
\vspace{-33pt}
\begin{IEEEbiography}[{\includegraphics[width=1in,height=1.25in,clip,keepaspectratio]{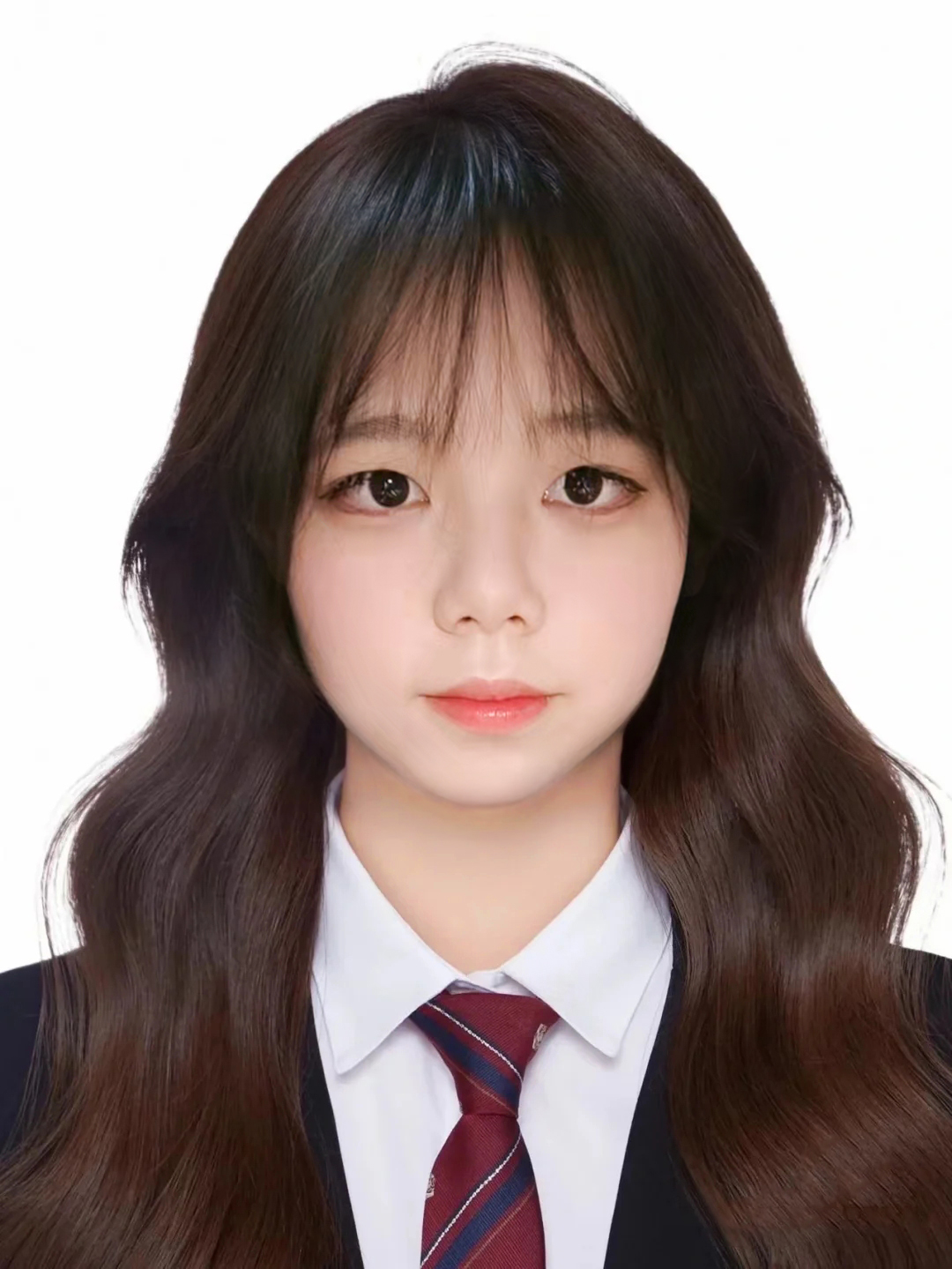}}]{Yuwei Li}
received the B.S. degree in public administration from Huazhong University of Science and Technology, Wuhan, China, 2022. She is currently working towards master's degree with the School of Public Administration, Renmin University of China, Beijing. Her current research interests include word representations, deep learning.
\end{IEEEbiography}

\vspace{11pt}
\vspace{-33pt}
\begin{IEEEbiography}[{\includegraphics[width=1in,height=1.25in,clip,keepaspectratio]{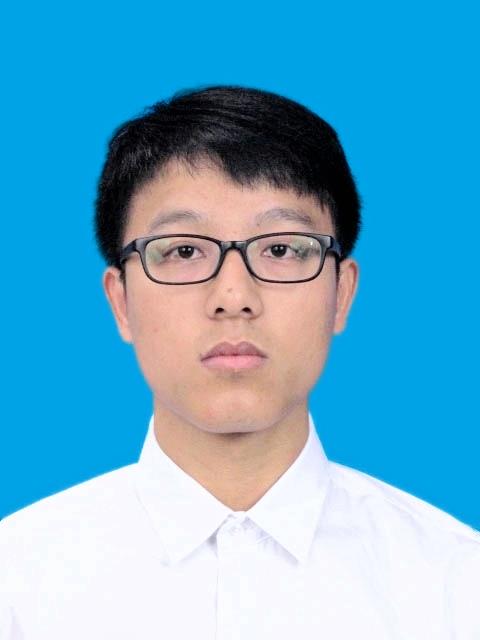}}]{Weifeng Li}
received the B.S. degree in mechanical manufacturing and automation from the Qingdao University of Science and Technology, Qingdao, China, in 2018. He is currently working toward the M.S. degree with the State Key Laboratory of Digital Manufacturing Equipment and Technology, Huazhong University of
Science and Technology, Wuhan, China. His current research interests include intelligent manufacturing and image processing.
\end{IEEEbiography}

\vspace{11pt}
\vspace{-33pt}
\begin{IEEEbiography}[{\includegraphics[width=1in,height=1.25in,clip,keepaspectratio]{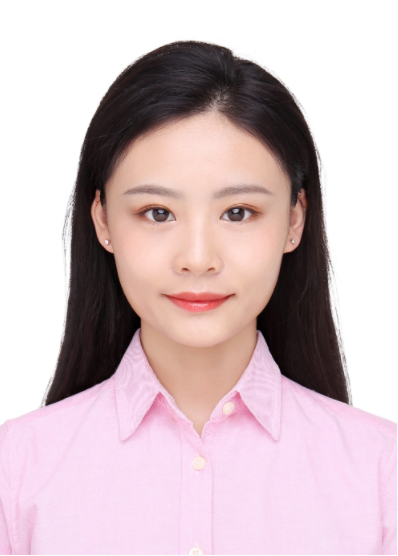}}]{Zhenrong Wang}
received the Master degree in Control Engineering from Harbin Institute of Technology in 2020. She is currently pursuing the Ph.D. at Huazhong University of Science and Technology, Wuhan. Her research
interest covers intelligent manufacturing, defect detection, machine learning and image processing.
\end{IEEEbiography}

\vfill

\end{document}